\documentclass{article}
\usepackage{colm2024_conference}

\usepackage{booktabs}
\usepackage{graphicx}
\usepackage{microtype}
\usepackage{amsmath,amsfonts}
\usepackage{enumitem}
\usepackage{caption}
\usepackage{subcaption}
\usepackage{xcolor}
\usepackage{multirow}
\usepackage{tabularx}
\usepackage{adjustbox}
\usepackage{float}
\usepackage{url}
\usepackage{hyperref}

\usepackage{amsmath,amsfonts,bm}

\def\eqref#1{equation~\ref{#1}}

\def\1{\bm{1}}

\DeclareMathAlphabet{\mathsfit}{\encodingdefault}{\sfdefault}{m}{sl}
\SetMathAlphabet{\mathsfit}{bold}{\encodingdefault}{\sfdefault}{bx}{n}

\newcommand{\E}{\mathbb{E}}

\newcommand{\avtrace}{\textsc{AVTrace}}

\newcommand{\gemmatemporal}{Gemma4-E4B-Temporal}
\newcommand{\anafig}[1]{files/#1}
\newcommand{\reviewhl}[1]{#1}
\definecolor{reviewpurple}{RGB}{231,211,255}

\title{\avtrace: Diagnosing Audio-Visual Temporal \\Reasoning in Omni Models}

\author{
  \vspace{4pt}
  \textbf{Longyin Zhang}$^{1}$, \textbf{Parth Sakhare Mahendra}$^{1}$, \textbf{Chengwei Wei}$^{1,2}$, \\
  \textbf{Ning Zhang}$^{1}$, \textbf{Lim Ming Chong}$^{1}$, \textbf{Sirui He}$^{1}$, and \textbf{Ai Ti Aw}$^{1}$ \\
  \vspace{0.5em}
  \begin{tabular}{c}
    \noalign{\vspace{0.2em}}
    $^{1}$Institute of Advanced Intelligence and Computing (IAIC), A*STAR, Singapore \\
    $^{2}$Centre for Frontier AI Research (CFAR), A*STAR, Singapore \\
    \noalign{\vspace{0.2em}}
    \texttt{Zhang\_Longyin@a-star.edu.sg}
  \end{tabular}
}

\begin{document}
\maketitle

\begin{abstract}
Omni models can describe video content, but can they locate events in time, preserve event order, and judge audio-visual synchronization? We introduce {\avtrace} (Audio-Visual Temporal Reasoning Assessment and Capability Evaluation), a silver-standard diagnostic suite spanning onset and span grounding, synchronization, next-step prediction, cross-modal localization, chain parsing, and event-conditioned comprehension. It contains 34,114 training examples and category-balanced development and test splits of 3,500 and 7,000 examples. We evaluate five open omni models under their respective input configurations using reference-blind response normalization followed by deterministic scoring. All five off-the-shelf systems score below the test split's majority-label baseline of 0.556 on synchronization verification, and obtain low scores on chain parsing and event-conditioned grounding and comprehension. Development-set perturbations reveal task-dependent sensitivity in Qwen3-Omni-30B to modality removal and changes in visual input processing, without isolating their underlying causes. Parameter-efficient temporal post-training improves Gemma4-E4B-it on several benchmark metrics. On three external image benchmarks, task metrics change modestly, including some degradations, while teacher-forcing perplexity decreases. Together, these findings show that \textbf{semantic reference-text overlap should not be treated as a proxy for temporal localization}, and that \avtrace{} can identify task-specific weaknesses while providing a testbed for temporal post-training.
\end{abstract}

\section{Introduction}
\label{sec:intro}

Recent omni models support joint interaction over video, audio, and language~\citep{xu2025qwen25omni,xu2025qwen3omni,minicpmo45card2025}. Standard video QA scores, however, need not isolate whether a system represents \emph{when} evidence
occurs or whether evidence from different streams is aligned~\citep{liu2024tempcompass,li2024vitatecs,lu2026fave}. Diagnostic benchmarks
for multimodal video understanding and cross-modal temporal alignment are therefore increasingly used to evaluate these questions explicitly~\citep{patraucean2023perception,liu2024tempcompass,li2024vitatecs,lu2026fave,zhong2026havebench,zhou2025dailyomni}.
Audio-visual event localization and synchronization also have a longer history as learning tasks, providing the task foundations on which these diagnostics build~\citep{tian2018ave,korbar2018cooperative}. Distinguishing content recognition from temporal grounding matters for applications such as incident review, dubbing QA, assistive agents, and procedural understanding.

The seven categories in Table~\ref{tab:tasks} probe distinct forms of temporal competence. Existing systems differ substantially in temporal sampling, media preprocessing, and multimodal fusion interfaces~\citep{xu2025qwen25omni,xu2025qwen3omni,minicpmo45card2025,gemma4card2026,interactiveomnicard2025}. This makes content recognition an insufficient proxy for binding evidence to the physical timeline, preserving its order, or comparing it against another modality at fine temporal resolution~\citep{tian2018ave,zhou2025dailyomni,zhu2025ravst,liu2024tempcompass,li2024vitatecs,lu2026fave}.

We study this question with {\avtrace}, a diagnostic benchmark assembled from existing audio-video sources and rendered task clips. Rather than claiming a single universal ``temporal reasoning'' score, the benchmark decomposes the problem into \textbf{seven categories}: Cat1 (Minimal-span temporal grounding), Cat2 (Event-span temporal grounding), Cat3 (A/V synchronization verification), Cat4 (Next-step identification), Cat5 (Conditional cross-modal localization), Cat6 (Temporal chain parsing), and Cat7 (Event-conditioned temporal grounding and comprehension). This decomposition lets us distinguish a model that can describe an event from one that can localize it or preserve its order. Its relationship to prior diagnostic video and audio-visual benchmarks, which cover broad perceptual skills, event localization, cross-modal temporal alignment, and hierarchical audio-visual evaluation~\citep{patraucean2023perception,tian2018ave,zhou2025dailyomni,lu2026fave,zhong2026havebench}, is discussed in Section~\ref{sec:related}.

Our study targets \textbf{observable system behavior rather than internal model mechanisms}: we evaluate end-to-end behavior and do not infer internal representations or architectural mechanisms from benchmark scores. We separate final held-out evaluation from development-set interventions: test examples are used only for final model comparison, whereas all perturbation design and analysis use the disjoint development split. Held-out results characterize each system's performance under a fixed protocol, while controlled interventions on the development partition measure how sensitive that performance is to the evidence the tasks demand.

Our contributions are:
\begin{itemize}[leftmargin=0.4cm]
  \item We present {\avtrace}, a structured seven-task diagnostic suite, with a 7,000-item category-balanced test set reserved for final evaluation and excluded from model development and perturbation analyses, and a separate 3,500-item category-balanced development set for perturbation analysis.
  \item We propose an evaluation package, including prediction-level records, normalized outputs, and a deterministic scoring implementation, to support reproducible comparisons under the reported configurations and fine-grained analysis of model behavior.
  \item We report benchmark results for five models on the test split, and paired intervention analyses on the development split that diagnose how scores change when evidence is removed or perturbed. We further show that parameter-efficient temporal post-training can substantially improve a small omni model: at least one post-trained Gemma4-E4B variant exceeds all off-the-shelf systems on the headline metrics for Cat1, Cat3, Cat4, and Cat7, while exhibiting task-dependent performance changes and lower perplexity on three external image benchmarks and five additional audio-task subsets.
\end{itemize}

\section{\avtrace{} Benchmark Construction}
\label{sec:benchmark}
\avtrace{} contains 44,614 silver-standard audio-video examples across the seven categories in Table~\ref{tab:tasks}. The categories isolate complementary temporal demands, from minimal event onsets and event spans to synchronization, prediction, cross-modal localization, ordered procedures, and event-conditioned comprehension. Their final references are not all derived in the same way: depending on the category, they combine upstream annotations, deterministic transformations, controlled rendering, and model-assisted construction.
\begin{table}[t]
\centering
\small
\caption{The seven \avtrace{} diagnostic categories.}
\label{tab:tasks}
\begin{tabular}{m{.06\linewidth}m{.54\linewidth}m{.24\linewidth}}
\toprule
Cat. & Capability & Output \\
\midrule
Cat1 & Minimal-span temporal grounding & Timestamp \\
Cat2 & Event-span temporal grounding & Start/end span(s) \\
Cat3 & A/V synchronization verification & Yes/no \\
Cat4 & Next-step identification & Next action \\
Cat5 & Conditional cross-modal localization & Span + description \\
Cat6 & Temporal chain parsing & Ordered steps \\
Cat7 & Event-conditioned temporal grounding and comprehension & Anchor, target, answer \\
\bottomrule
\end{tabular}
\end{table}



\subsection{Source-to-target conversion}
We construct \avtrace{} from public source datasets with complementary temporal signals. Appendix~A (Table~\ref{tab:provenance}) summarizes the upstream supervision, model-assisted construction stages, and deterministic validation gates for each category. Upstream labels are treated as task-specific supervision signals rather than copied unchanged into the final benchmark. The conversion paths distinguish direct temporal supervision from weak anchors, synthetic labels, and model-assisted references, which have different implications for reference quality:
\begin{itemize}[leftmargin=0.4cm]
  \item \textbf{Cat1:} Event or sound spans and timestamps become onset-grounding questions; model-assisted temporal refinement\footnote{``Model-assisted'' denotes silver-reference construction stages using Qwen3-Omni-30B-A3B-Instruct~\citep{xu2025qwen3omni} for audio-visual grounding and Qwen3-235B-A22B-Instruct-2507~\citep{qwen3card2025} for question generation and text-based validation.} supplies the final timestamp when the source annotation is only a coarse cue:
  Qwen3-Omni watches and listens to the entire clip and is asked when the source-specified sound first becomes audible, returning a single timestamp in seconds; responses that cannot be parsed or that fall outside the clip are rejected. For Perception Test items, whose source labels are coarse, the clip is first summarized and the label is rewritten into a refined question before the timestamp answer is elicited. We reject onsets within 0.1 seconds of the clip start, and for onsets within the first three seconds we prepend neutral context---an event-free clip drawn from a shared pool sampled across sources and filtered to exclude speech and singing, so it cannot leak the queried event.
  \item \textbf{Cat2:} Event and action intervals become candidate span-grounding references. For ambiguous AVE and LLP/AVVP~\citep{tian2020avvp} labels, Qwen3-Omni selects the audio- or visual-focused question variant whose predicted span best overlaps the source interval, without revising its boundary. Qwen3-Omni then provides full-clip and span-specific observations for Qwen3-235B to classify each candidate as \emph{present}, \emph{partial}, or \emph{absent}. We retain present and partial spans with their original boundaries, discard records with no remaining spans, and pad spans within one second of a clip boundary with neutral context.
  \item \textbf{Cat3:} Original synchronized media from ten public sources, including MUSIC-AVQA~\citep{li2022musicavqa}, WASD~\citep{roxo2025wasd}, AVSBench~\citep{zhou2022avsbench}, and UnAV-100~\citep{geng2023unav100} (Appendix~A, Table~\ref{tab:source-provenance}), supply positive examples; desynchronized negatives are rendered with one of four mechanisms: a global shift of the audio track, a local shift of an event region, local audio replacement, or cross-splice rendering. An audio-visual speech probe (Qwen3-Omni) labels each source video for the presence of audible speech; as a conservative construction choice, shift-based negatives are discarded for such videos, which contribute synchronization positives and, for a subset, content-replacement negatives.
  \item \textbf{Cat4:} Ordered, temporally annotated action sequences from COIN~\citep{tang2019coin}, Perception Test~\citep{patraucean2023perception}, and EPIC-KITCHENS-100~\citep{damen2022epic} are rendered into a task clip that spans three consecutive observed steps together with the immediately following target action, padded by two seconds of context on each side. The question names the three observed steps and asks which action comes next. At inference, every system instead receives a cached prefix ending at $\max(\min(e_{\mathrm{obs}}+2\,\mathrm{s}, s_{\mathrm{target}}-1\,\mathrm{s}), e_{\mathrm{obs}})$, where $e_{\mathrm{obs}}$ is the maximum end time among the observed steps and $s_{\mathrm{target}}$ is the annotated target-action start time. The prefix is designed to withhold the target action while retaining the observed sequence. The task therefore measures anticipation of the next action from the observed sequence.
  \item \textbf{Cat5:} Each example conditions on an event in one modality and asks for the corresponding event in the other: given an audible event, the model must localize the temporally corresponding visible event and describe it (audio-to-visual), and given a visible event, it must localize and describe the corresponding sound (visual-to-audio). Source event spans serve only as candidate anchors: Qwen3-Omni identifies an event that occurs exactly once in the clip, specifies a short temporal window around it, and describes its visual and auditory content. Qwen3-235B then generates a question from this annotation and verifies that answering it requires cross-modal reasoning while yielding a valid answer in the target modality. Finalization rejects abstentions, spans that fall outside the clip or do not overlap the paired audio-visual event, spans covering 60\% or more of the clip duration to discourage overly broad localization predictions, and answers that do not match the requested target-modality schema.
  \item \textbf{Cat6:} We split instructional videos and narration sequences into short clips, with each step labeled by its start and end time within the clip. Qwen3-Omni then checks each step using the step itself and its neighboring steps. For datasets with standard step labels, it checks only the step interval. The model may confirm the step, correct its label or timing, or reject it. We discard any record containing an unverified step. Finalization then exports the answer deterministically from the verified chain and requires at least two positive-duration steps with strictly increasing start times; model assistance may refine individual step labels and boundaries but cannot add or reorder steps in the released reference.
  \item \textbf{Cat7:} Each example specifies an anchor event, a temporal relation between the anchor and a target interval, and a free-text answer about the target. Event spans, chains, and event-centered questions provide anchor candidates, which Qwen3-Omni refines or relabels on local clips. After refinement, an
  annotated neighboring action is used as the target only when it begins or ends within three seconds of the anchor; otherwise, a local window of three to ten seconds, scaled to the anchor's duration, is constructed adjacent to the anchor. Context clips are re-encoded while preserving the intended temporal boundaries, and finalization requires a valid anchor, target, and answer whose temporal geometry is consistent with the query.
\end{itemize}

\begin{table}[t]
\centering
\small
\setlength{\tabcolsep}{2pt}
\caption{Statistics of the 44,614-record pool. Durations are the median and 95th-percentile released clip lengths, respectively.}
\label{tab:pool-statistics}
\begin{tabular}{m{.07\textwidth}m{.08\textwidth}m{.14\textwidth}m{.66\textwidth}}
\toprule
Cat. & $N$ & Dur. (s) & Structure \\
\midrule
Cat1 & 6,129 & 28.3 / 195.7 & One onset timestamp. \\
Cat2 & 3,720 & 110.6 / 230.1 & 3,423 single-span and 297 multi-span records. \\
Cat3 & 6,638 & 50.0 / 550.7 & 3,319 synchronized and 3,319 desynchronized records across four rendering mechanisms. \\
Cat4 & 9,992 & 20.6 / 100.0 & Every record has three observed steps and one target step. \\
Cat5 & 6,943 & 23.2 / 118.3 & Audio-to-visual description: 4,210; visual-to-audio description: 2,733. \\
Cat6 & 3,614 & 68.2 / 205.8 & Chain length: median 3 steps, P95 14 steps. \\
Cat7 & 7,578 & 64.8 / 459.6 & Query relation: after 4,611; before 2,964; during 3 (residual). \\
\bottomrule
\end{tabular}
\end{table}

\subsection{Human-in-the-loop refinement}
We use human review as a diagnostic development loop rather than as release-wide annotation. In each review-and-rebuild iteration, reviewers inspect 32 records per category (224 in total), sampled to cover upstream source datasets as evenly as possible, and record task validity, reference correctness, error categories, and free-text notes. Recurring failure modes are translated into revisions of source converters, prompts, and post-processing rules, after which the affected category is rebuilt from the current usable pool. We conducted 18 such versioned iterations. This process is intended to identify and remove systematic construction failures.

Before release, we apply deterministic validity checks for media availability, schema compliance, temporal bounds, and category-specific structural constraints such as ordered steps or query-consistent spans. Records with unresolved construction failures, abstentions, or invalid outputs are excluded. These gates ensure release-level structural and media validity, but do not independently verify the semantic correctness of every source or model-assisted reference.

\subsection{Fixed partitions and benchmark scope}
\label{sec:partitions}
The resulting usable pool is partitioned into 34,114 training, 3,500 development, and 7,000 test examples, with 500 and 1,000 examples per category in the development and test splits, respectively. To preserve provenance isolation, we group examples by the pair (upstream dataset, original video ID) and assign each dataset-qualified source-video group to a single split. This dataset-qualified grouping can constrain exact label balancing within each split. For example, Cat3's test partition contains 444 synchronized and 556 desynchronized examples (majority-label baseline 0.556), which we report when interpreting Cat3 accuracy. More broadly, the benchmark is \textbf{silver-standard rather than human-verified gold data}: it supports behavioral comparison on this fixed release, rather than estimates of human agreement or reference correctness, and we do not estimate same-family model bias in model-assisted references.

\section{Evaluation Protocol}
\label{sec:protocol}

\subsection{Evaluated systems and input interfaces}
We evaluate Qwen3-Omni-30B~\citep{xu2025qwen3omni}, Qwen2.5-Omni-3B \citep{xu2025qwen25omni}, MiniCPM-o-4.5~\citep{minicpmo45card2025}, Gemma4-E4B-it~\citep{gemma4card2026}, and InteractiveOmni-4B \citep{interactiveomnicard2025}. We evaluate each system through its supported input interface (Appendix~\ref{app:systems}). For Cat4, systems receive a derived prefix that excludes the target action, as defined in Section~\ref{sec:benchmark}; for all other categories, systems receive the full released clip. The interfaces differ in ways that matter for temporal tasks. Qwen3-Omni-30B and MiniCPM-o-4.5 receive the full supplied clip. Qwen2.5-Omni-3B samples video uniformly over the supplied clip with at most 300 frames and receives audio with a 300-second feature window. Gemma4-E4B-it and InteractiveOmni-4B receive only the first 30 seconds of video and audio; references beyond that window are unavailable to those systems. Cross-model comparisons on this release therefore characterize end-to-end systems under their supported interfaces rather than architectures evaluated under matched media budgets or temporal coverage.

\subsection{Output normalization and deterministic scoring}
Models are instructed to emit the compact structured outputs in Table~\ref{tab:tasks}. We first apply deterministic schema parsing as a separate diagnostic branch: it extracts the first well-formed JSON value from each response, permits incidental surrounding prose and Markdown code-block delimiters, and applies category-specific validation and scoring without interpreting free-text answers or repairing malformed JSON. The extracted value is checked against the category schema: a numeric onset in seconds (Cat1); one span object or a JSON list of span objects (Cat2); a yes/no string (Cat3); an action string or an object with an \texttt{action} field (Cat4); a JSON object with keys \texttt{start}, \texttt{end}, and \texttt{description} (Cat5); a JSON list whose object elements are interpreted as step objects in output order (Cat6); and a JSON object with keys \texttt{anchor\_start}, \texttt{anchor\_end}, \texttt{target\_start}, \texttt{target\_end}, and \texttt{answer} (Cat7). Parser validity is reported separately from the normalized headline results.

A reference-blind \reviewhl{Qwen3.5-27B} normalizer~\reviewhl{\citep{qwen35card2026}} then receives only the question, video duration, and raw response; \textbf{it does not receive the reference answer}. With a category-specific prompt, it extracts the intended structured answer from the raw response, whether the response wraps the answer in JSON or markdown or states times and steps in free text (e.g., ``between 10 and 20 seconds''), and returns an ok/invalid status; deterministic task-specific scoring is then applied to the extracted prediction. This step is therefore model-based response parsing rather than format-only cleaning: it can recover answers that the strict parser rejects, but its extraction decisions have not been independently validated against human annotations.

Scoring follows the target structures rather than a shared benchmark-wide metric. The metrics are composed from a small set of shared primitives (i.e., temporal overlap, token overlap, exact match, and thresholded accuracy) with task-specific matching and aggregation rules. Each category reports the metrics applicable to its target structure.

Thresholded timestamp accuracy and mean absolute error, for a predicted onset $\hat{t}$ against reference time $t$, are
\begin{align}
\mathrm{Acc}@\tau&=\mathbb{1}\left[\left|\hat{t}-t\right|\le\tau\right],\\
\mathrm{MAE}&=\frac{1}{n}\sum_{i=1}^{n}\left|\hat{t}_i-t_i\right|.
\end{align}
Invalid predictions enter these two statistics differently: $\mathrm{Acc}@\tau$ is averaged over all examples, with an invalid or missing onset counted as incorrect, whereas MAE is computed only over examples that produce a valid numeric onset. The valid-output rate is reported alongside both statistics, so the coverage on which MAE is conditioned is visible and the two denominators can be compared directly.

Let $a=[s_a,e_a]$ and $b=[s_b,e_b]$ be reference and predicted spans, with $s_a\le e_a$ and $s_b\le e_b$. The temporal intersection-over-union is
\begin{equation}
\operatorname{tIoU}(a,b)=\frac{\max\left(0,\,\min(e_a,e_b)-\max(s_a,s_b)\right)}
{\max(e_a,e_b)-\min(s_a,s_b)}.
\end{equation}
We allow zero-duration spans with $s_a=e_a$; the denominator is zero only when both intervals coincide at the same point, in which case we define $\operatorname{tIoU}=1$. Predicted spans whose bounds are non-numeric or whose start exceeds their end fail schema validation and are scored as task failures; all other predictions, including spans outside the clip, are scored without modification.

For reference spans $R$ and predicted spans $P$, let $m_\tau$ be the maximum number of one-to-one span pairs satisfying $\operatorname{tIoU}\ge\tau$. Such maximum-cardinality matchings need not be unique; we compute $m_\tau$ with a deterministic augmenting-path procedure that considers each reference span's eligible predictions in descending order of $\operatorname{tIoU}$ (duplicate predictions are treated as distinct candidates), and let $M_{0.5}$ denote the matching returned at $\tau=0.5$. The span-level score and the mean overlap of matched pairs are
\begin{equation}
F_1@\tau=\frac{2m_\tau}{|R|+|P|},
\end{equation}
\begin{equation}
\overline{\operatorname{tIoU}}=\frac{1}{|M_{0.5}|}\sum_{(a,b)\in M_{0.5}}
\operatorname{tIoU}(a,b).
\end{equation}
When $M_{0.5}$ is empty, we set $\overline{\operatorname{tIoU}}=0$, so an example whose predicted spans match no reference span above threshold contributes the minimum value rather than being excluded from the average.

For text, let $r$ and $p$ be the tokenized reference and prediction, and let $h$ be the size of their multiset overlap. Both strings are first normalized with Unicode NFKC\footnote{Unicode Normalization Form Compatibility Composition (NFKC) maps compatibility equivalents to a common representation, for example, a full-width Latin capital A to A and an fi ligature to fi. This prevents compatibility-equivalent Unicode forms from affecting the score.} and lowercased, and tokens are the maximal contiguous sequences of Unicode word characters, so whitespace and punctuation carry no weight. If one side yields no tokens, the text scores are defined as $1$ when both sides are empty and $0$ otherwise. Token F1 and ROUGE-L~\citep{lin2004rouge} are
\begin{align}
F_{\mathrm{tok}}&=\frac{2h}{|r|+|p|},\\
R_{\mathrm{L}}&=\frac{2\,\mathrm{LCS}(r,p)}{|r|+|p|},
\end{align}
where $\mathrm{LCS}(r,p)$ is the length of the longest common subsequence over these
token sequences; exact match additionally compares the fully token-normalized
strings.

Joint scores combine temporal overlap with text quality. The Cat5 single-span
localization score and the Cat7 anchor--target grounding score are
\begin{align}
J_{\mathrm{loc}}&=\sqrt{\operatorname{tIoU}(a,b)\cdot F_{\mathrm{tok}}},\\
J_{\mathrm{at}}&=\sqrt{\operatorname{tIoU}(a_A,b_A)\cdot\operatorname{tIoU}(a_T,b_T)}
\cdot F_{\mathrm{tok}},
\end{align}
where subscripts $A$ and $T$ denote anchor and target spans and $F_{\mathrm{tok}}$ is computed on the accompanying description or answer text.
For chain parsing, let $R$ and $P$ denote the reference and predicted step sequences, respectively. Each step is represented as a span-text pair and matched under
\begin{equation}
\mathrm{hit}(i,j)=\mathbb{1}\left[\operatorname{tIoU}(a_i,b_j)\ge 0.5\;\wedge\;
F_{\mathrm{tok}}(u_i,v_j)\ge 0.5\right],
\end{equation}
where $u_i$ and $v_j$ are the reference and predicted step texts, and predicted steps are taken in the model's output order. Ordered matching computes $m_{\mathrm{ord}}$, the maximum number of hits whose indices are strictly increasing in both the reference sequence and the predicted sequence, obtained by the standard longest-common-subsequence dynamic program over the hit indicators. Unordered matching computes the maximum cardinality of a one-to-one assignment between reference and predicted steps restricted to hits. Each StepF1 score takes the form $2m/(|R|+|P|)$ using the match count $m$ from its respective matching procedure. A matching procedure may admit multiple equally large maximum matchings that pair different steps. Since the score uses only their shared cardinality, rather than the identities of the paired steps, its value is unaffected by how such ties are resolved.

Applied to the seven categories: Cat1 reports $\mathrm{Acc}@\tau$ for $\tau\in\{0.5,1\}$ and MAE; Cat2 reports $F_1@\tau$ for $\tau\in\{0.3,0.5,0.7\}$ and $\overline{\operatorname{tIoU}}$ over multi-span outputs; Cat3 reports the accuracy $\mathrm{Acc}=\mathbb{1}[\hat{y}=y]$ of binary synchronization labels; Cat4 reports $F_{\mathrm{tok}}$, $R_{\mathrm{L}}$, and exact match for the next-action text; Cat5 reports $\operatorname{tIoU}$, the text metrics, and the single-span joint score $J_{\mathrm{loc}}$; Cat6 reports ordered and unordered StepF1 together with exact-chain accuracy $\mathbb{1}[m_{\mathrm{ord}}=|R|=|P|]$; Cat7 reports anchor and target $\operatorname{tIoU}$, the answer text metrics, and the anchor--target joint score $J_{\mathrm{at}}$. Except for Cat1 MAE, every reported performance metric is averaged over all examples in its category; invalid normalized predictions receive zero for every defined metric.

\subsection{Development-set behavioral probes}
All perturbation experiments use the 3,500-item development split and pair each condition with the same system under the full-input baseline. The full-input baseline is exactly the main protocol of Section~\ref{sec:protocol}: each system receives the media interface specified in Appendix~\ref{app:systems}. For example, Qwen3-Omni-30B receives native video configured at 1\,FPS together with the full audio track, with greedy decoding and a 512-token budget. The conditions are as follows:
\begin{itemize}[leftmargin=0.4cm]
  \item \textbf{Full input:} video, audio, and the task question are supplied once with the schema-constrained task prompt.
  \item \textbf{Audio only:} the clip's full audio track, extracted as 16\,kHz mono WAV, and the task question are supplied; no video input is provided.
  \item \textbf{Vision only:} Qwen3-Omni-30B receives the same native video configured at 1\,FPS as in the full-input condition, together with the task question, but audio is disabled.
  \item \textbf{Text only:} only the task question and the task prompt are supplied.
  \item \textbf{Reduced visual sampling:} the full condition is retained, but Qwen3-Omni-30B receives video configured at 0.125\,FPS.
  \item \textbf{Temporally permuted frames:} the task inputs are retained, but we supply a video re-encoded from 1\,FPS-extracted JPEG frames after randomly reordering those frames with a fixed seed; the corresponding audio is included in the re-encoded video. This condition combines resampling/re-encoding with a frame-order perturbation rather than providing a pure frame-permutation control. The reported effect is conditional on this fixed-seed run.
  \item \textbf{Two-stage prompting:} the system first summarizes the media in a few sentences, covering salient visual and audible events and their apparent order, without answering the task. A fresh second call then receives the same media, the generated summary, and the original schema-constrained task prompt; only this response is scored. Both calls use greedy decoding with a 512-token budget, and the media is re-processed for the second call.
\end{itemize}
 
For each category and perturbation, we pair the perturbation result for each item with that same system's full-input result on the same item, and report paired-bootstrap 95\% intervals (10,000 resamples) for the mean difference in the category-specific primary metric~\citep{efron1993bootstrap}. All records available in both conditions are retained, and invalid normalized predictions receive a zero on the defined primary metric.

\section{Main Benchmark Results}
\label{sec:results}

Table~\ref{tab:test-leaderboard} reports component metrics on the 7,000-item test split after reference-blind normalization and deterministic scoring. Among the off-the-shelf systems, Qwen3-Omni-30B has the highest value for most test metrics, whereas InteractiveOmni-4B is highest on Cat3 accuracy. At least one post-trained \gemmatemporal{} variant (described below) exceeds every off-the-shelf system on the headline metrics for Cat1, Cat3, Cat4, and Cat7. The category-specific results reveal different failure profiles across point grounding, span grounding, synchronization, prediction, cross-modal localization, chain recovery, and event-conditioned comprehension.

\paragraph{Post-trained Gemma4 references.} The two \gemmatemporal{} columns are post-trained variants included as domain-adapted reference points. The labels \emph{Gold} and \emph{Matched} denote training-target variants, not reference quality: \emph{Gold} uses temporal-only canonical silver-reference targets, including Cat4's full observed-step and target-step timeline. \emph{Matched} uses temporal-only targets formed by reference-conditioned Qwen3-235B correction of base-model outputs, with fallback to the canonical target when correction fails. We initialize from the released Gemma 4 E4B instruction-tuned checkpoint (\path{google/gemma-4-E4B-it}), and train a LoRA adapter~\citep{hu2022lora} (rank 64, $\alpha=128$, dropout 0.1). Adapters are attached to all decoder attention and MLP projections ($q$, $k$, $v$, $o$, gate, up, down) and to the vision and audio modality projectors; the vision and audio encoder towers, the output head (\texttt{lm\_head}), and the per-layer embedding and projection modules are frozen, so only the LoRA parameters (140M, 1.7\% of the total) are updated. Training records use category-specific preprocessing: non-Cat4 temporal records are cropped and re-based as needed, whereas Cat4 uses a target-withholding prefix; audio is limited to the first 30\,s in all cases. Training uses the instruction-tuned chat template with 30 sampled frames. Optimization uses AdamW with learning rate $2\times10^{-5}$ and linear decay to zero without warmup, an effective batch size of 256, approximately four epochs (536 optimizer steps), seed 42, and bf16 with gradient checkpointing.

\begin{table*}[t]
\centering
\small
\setlength{\tabcolsep}{2pt}
\renewcommand{\arraystretch}{1.05}
\caption{Test-set results under \reviewhl{reference-blind normalization} and deterministic scoring ($n=1{,}000$ per category). Values are category-level means; invalid normalized predictions receive zero for every defined metric. Cat1 MAE is the exception and averages only valid numeric timestamps. Valid output rate denotes scoreable normalized responses. Bold marks the best result in each row across all systems.}
\label{tab:test-leaderboard}
\begin{tabular}{@{}lccccc@{\hspace{8pt}}cc@{}}
\toprule
& \multicolumn{5}{c@{\hspace{8pt}}}{Off-the-shelf systems} & \multicolumn{2}{c}{Post-trained refs} \\
\cmidrule(lr){2-6}
\cmidrule(l){7-8}
 Category / metric & \shortstack{Qwen3-\\Omni-30B} & \shortstack{MiniCPM-o-\\4.5} & \shortstack{Qwen2.5-\\Omni-3B} & \shortstack{Interactive\\Omni-4B} & \shortstack{Gemma4-\\E4B-it} & \shortstack{Temporal\\(\emph{Gold})} & \shortstack{Temporal\\(\emph{Matched})} \\
\midrule
\multicolumn{8}{@{}l}{\textbf{Cat1: Minimal-span temporal grounding}} \\
 \hspace{1em}Valid output rate $\uparrow$ & \textbf{1.000} & 0.971 & 0.998 & \textbf{1.000} & \reviewhl{0.986} & \textbf{1.000} & \textbf{1.000} \\
 \hspace{1em}Acc@0.5s $\uparrow$ & 0.253 & 0.066 & 0.086 & 0.039 & 0.150 & \textbf{0.259} & \textbf{0.259} \\
 \hspace{1em}Acc@1s $\uparrow$ & 0.426 & 0.169 & 0.156 & 0.053 & 0.273 & 0.440 & \textbf{0.450} \\
 \hspace{1em}MAE (s) $\downarrow$ & 3.185 & 3.848 & 4.822 & 16.993 & \reviewhl{6.265} & 3.046 & \textbf{2.984} \\
\midrule
\multicolumn{8}{@{}l}{\textbf{Cat2: Event-span temporal grounding}} \\
 \hspace{1em}Valid output rate $\uparrow$ & \reviewhl{0.967} & \reviewhl{0.995} & \reviewhl{0.980} & \reviewhl{\textbf{1.000}} & \textbf{1.000} & \textbf{1.000} & \textbf{1.000} \\
 \hspace{1em}F1@tIoU.3 $\uparrow$ & \reviewhl{\textbf{0.402}} & \reviewhl{0.154} & 0.071 & 0.033 & \reviewhl{0.056} & 0.177 & 0.188 \\
 \hspace{1em}F1@tIoU.5 $\uparrow$ & \reviewhl{\textbf{0.265}} & \reviewhl{0.091} & \reviewhl{0.035} & 0.010 & \reviewhl{0.026} & 0.103 & 0.121 \\
 \hspace{1em}F1@tIoU.7 $\uparrow$ & \textbf{0.130} & 0.038 & 0.009 & 0.001 & 0.011 & 0.055 & 0.067\\
 \hspace{1em}Mean matched tIoU $\uparrow$ & \reviewhl{\textbf{0.198}} & \reviewhl{0.064} & 0.032 & 0.006 & 0.019 & 0.077 & 0.092 \\
\midrule
\multicolumn{8}{@{}l}{\textbf{Cat3: A/V synchronization verification}} \\
 \hspace{1em}Valid output rate $\uparrow$ & 0.976 & 0.922 & 0.921 & \textbf{1.000} & \textbf{1.000} & \textbf{1.000} & \textbf{1.000} \\
 \hspace{1em}Accuracy $\uparrow$ & 0.435 & 0.414 & 0.414 & 0.497 & 0.469 & \textbf{0.576} & 0.571 \\
\midrule
\multicolumn{8}{@{}l}{\textbf{Cat4: Next-step identification}} \\
 \hspace{1em}Valid output rate $\uparrow$ & \textbf{1.000} & \reviewhl{0.837} & \textbf{1.000} & \reviewhl{0.994} & \reviewhl{0.982} & \textbf{1.000} & \textbf{1.000} \\
 \hspace{1em}Token F1 $\uparrow$ & 0.486 & \reviewhl{0.429} & \reviewhl{0.479} & \reviewhl{0.432} & \reviewhl{0.381} & \reviewhl{\textbf{0.539}} & 0.536 \\
 \hspace{1em}ROUGE-L $\uparrow$ & \reviewhl{0.483} & \reviewhl{0.425} & \reviewhl{0.475} & \reviewhl{0.429} & \reviewhl{0.379} & \textbf{0.534} & 0.532\\
 \hspace{1em}Exact match $\uparrow$ & 0.267 & \reviewhl{0.254} & \reviewhl{0.269} & \reviewhl{0.221} & \reviewhl{0.185} & \reviewhl{\textbf{0.307}} & \reviewhl{\textbf{0.307}}\\
\midrule
\multicolumn{8}{@{}l}{\textbf{Cat5: Conditional cross-modal localization}} \\
 \hspace{1em}Valid output rate $\uparrow$ & \textbf{1.000} & \reviewhl{0.994} & \reviewhl{0.999} & \reviewhl{0.977} & \reviewhl{\textbf{1.000}} & \textbf{1.000} & \textbf{1.000} \\
 \hspace{1em}tIoU $\uparrow$ & \textbf{0.415} & \reviewhl{0.227} & 0.227 & \reviewhl{0.121} & 0.187 & 0.358 & 0.382 \\
 \hspace{1em}Text F1 $\uparrow$ & \reviewhl{0.448} & 0.362 & \reviewhl{0.228} & \reviewhl{0.346} & 0.318 & \textbf{0.507} & 0.451 \\
 \hspace{1em}ROUGE-L $\uparrow$ & \reviewhl{0.377} & 0.317 & \reviewhl{0.203} & \reviewhl{0.295} & 0.271 & \textbf{0.427} & 0.387\\
 \hspace{1em}Joint score $\uparrow$ & \reviewhl{\textbf{0.373}} & \reviewhl{0.208} & 0.160 & \reviewhl{0.125} & 0.155 & 0.318 & 0.306\\
\midrule
\multicolumn{8}{@{}l}{\textbf{Cat6: Temporal chain parsing}} \\
 \hspace{1em}Valid output rate $\uparrow$ & \reviewhl{0.995} & \reviewhl{0.969} & \reviewhl{0.866} & 0.994 & \reviewhl{0.960} & \textbf{1.000} & \reviewhl{0.996} \\
 \hspace{1em}Ordered StepF1 $\uparrow$ & \textbf{0.065} & 0.017 & \reviewhl{0.042} & \reviewhl{0.023} & 0.003 & 0.063 & 0.041 \\
 \hspace{1em}Unordered StepF1 $\uparrow$ & \textbf{0.065} & 0.017 & \reviewhl{0.042} & \reviewhl{0.023} & 0.003 & 0.063 & 0.041\\
 \hspace{1em}Exact-chain accuracy $\uparrow$ & \textbf{0.003} & 0.000 & 0.000 & 0.000 & 0.000 & 0.001 & 0.000\\
\midrule
\multicolumn{8}{@{}l}{\textbf{Cat7: Event-conditioned temporal grounding and comprehension}} \\
 \hspace{1em}Valid output rate $\uparrow$ & \reviewhl{0.964} & \reviewhl{0.873} & \reviewhl{0.879} & 0.994 & \reviewhl{0.974} & \textbf{1.000} & \reviewhl{\textbf{1.000}} \\
 \hspace{1em}Anchor tIoU $\uparrow$ & \reviewhl{0.156} & \reviewhl{0.106} & \reviewhl{0.063} & \reviewhl{0.025} & 0.044 & 0.183 & \textbf{0.190} \\
 \hspace{1em}Target tIoU $\uparrow$ & \reviewhl{0.125} & \reviewhl{0.071} & \reviewhl{0.060} & \reviewhl{0.027} & \reviewhl{0.031} & \reviewhl{\textbf{0.197}} & \textbf{0.197}\\
 \hspace{1em}Answer Text F1 $\uparrow$ & \reviewhl{0.250} & \reviewhl{0.207} & \reviewhl{0.163} & \reviewhl{0.251} & \reviewhl{0.248} & \textbf{0.434} & 0.423\\
 \hspace{1em}Answer ROUGE-L $\uparrow$ & \reviewhl{0.214} & \reviewhl{0.176} & \reviewhl{0.144} & \reviewhl{0.216} & \reviewhl{0.208} & \textbf{0.356} & 0.352\\
 \hspace{1em}Joint score $\uparrow$ & \reviewhl{0.020} & \reviewhl{0.012} & 0.005 & 0.002 & 0.004 & 0.076 & \textbf{0.077}\\
\bottomrule
\end{tabular}
\end{table*}

\begin{figure*}[t]
\centering
\includegraphics[width=0.96\textwidth]{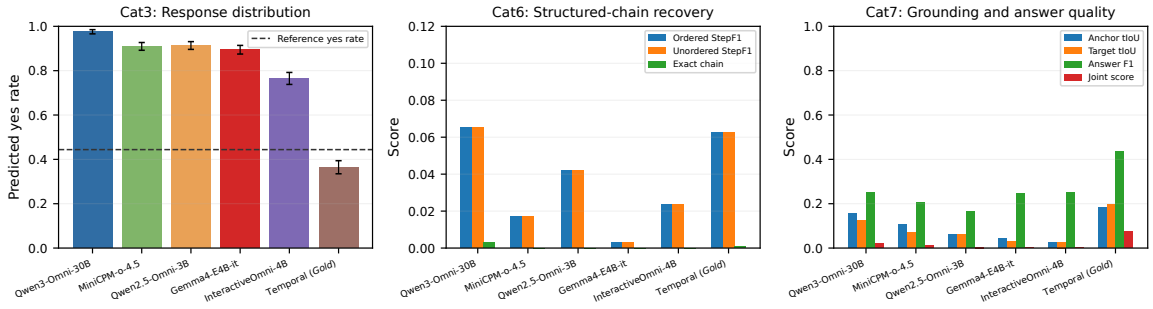}
\caption{Test component analysis ($n=1{,}000$ per category). Left: Cat3 predicted affirmative-response rates, with the reference positive-label rate shown as a dashed line; error bars are nonparametric bootstrap 95\% intervals. Center: Cat6 ordered and unordered step recovery and exact-chain accuracy. Right: Cat7 anchor and target localization with answer quality; invalid normalized predictions receive zero for all displayed Cat7 metrics.}
\label{fig:test-components}
\end{figure*}

We discuss the per-category results in Table~\ref{tab:test-leaderboard} in turn.
\begin{itemize}[leftmargin=0.4cm]
  \item \textbf{Minimal-span temporal grounding.} Both \gemmatemporal{} variants are highest at the 0.5\,s tolerance (0.259); \emph{Matched} is highest at 1\,s (0.450) and has the lowest MAE (2.984\,s), improving over the base Gemma4-E4B-it model (0.150, 0.273, and 6.265\,s) and over Qwen3-Omni-30B (0.253, 0.426, and 3.185\,s), the best off-the-shelf system. Tightening the tolerance from 1\,s to 0.5\,s retains only 39\%--74\% of each system's Acc@1s. The valid-output rates are high for all systems (0.971--1.000), so this pattern is not primarily explained by schema failure.
\item \textbf{Event-span temporal grounding.} Qwen3-Omni-30B leads at each tIoU threshold (\reviewhl{0.402, 0.265}, and 0.130 at 0.3, 0.5, and 0.7, respectively), but all systems decline as the required span overlap becomes stricter. The \emph{Matched} variant is second at every threshold (0.188, 0.121, and 0.067), improving over the base Gemma4-E4B-it model at every threshold, but remains well below Qwen3-Omni-30B. These results underscore the difficulty of recovering precise event spans.
\item \textbf{A/V synchronization verification} accuracy ranges from 0.414 to 0.576, with the \emph{Gold} variant highest at 0.576 (the base Gemma4-E4B-it model scores 0.469) and InteractiveOmni-4B highest among off-the-shelf systems at 0.497. The test partition contains 444 synchronized and 556 desynchronized examples, giving a majority-label baseline of 0.556; both post-trained variants modestly exceed this baseline. Figure~\ref{fig:test-components} (Left) shows that affirmative-response rates differ across systems, but this paper does not report results stratified by desynchronization strategy.
\item \textbf{Next-step identification.} The \emph{Gold} variant has the highest next-action reference-text overlap on Token F1 (\reviewhl{0.539}) and ROUGE-L (0.534), while \reviewhl{\emph{Gold} and \emph{Matched} tie for the highest exact match (0.307).} Among off-the-shelf systems, Qwen3-Omni-30B leads Token F1 and ROUGE-L (0.486 and \reviewhl{0.483}), while \reviewhl{Qwen2.5-Omni-3B has the highest exact match (0.269).} Both post-trained results are substantial gains over the base Gemma4-E4B-it model (\reviewhl{0.381, 0.379, and 0.185}).
\item \textbf{Conditional cross-modal localization.} Qwen3-Omni-30B leads on temporal overlap (0.415) and joint score (\reviewhl{0.373}), whereas the \emph{Gold} variant leads on description overlap (Text F1 0.507). The post-trained variants obtain tIoU values of 0.358--0.382 and joint scores of 0.306--0.318. Components remain separable for other systems: InteractiveOmni-4B reaches Text F1 \reviewhl{0.346}, near MiniCPM-o-4.5's 0.362, but has lower tIoU (\reviewhl{0.121} versus \reviewhl{0.227}) and joint score (\reviewhl{0.125} versus \reviewhl{0.208}). Descriptions can overlap substantially with the reference text while localizing the requested cross-modal event poorly in time.
\item \textbf{Temporal chain parsing.} Figure~\ref{fig:test-components} (Center) shows that ordered and unordered StepF1 are exactly equal for every evaluated model. For each system, 97\%--99.9\% of test records yield at most one content-matched step, and with zero or one matched step, the ordered and unordered match counts coincide by construction. \emph{Gold} nearly matches Qwen3-Omni-30B on both measures (0.063 versus 0.065), but \textbf{exact-chain accuracy remains near zero}: 0.003 for Qwen3-Omni-30B, 0.001 for \emph{Gold}, and 0 for others. Unmatched steps enter the F1 denominator and lower both scores equally, so the difference between the two scores reflects ordering only among steps that already matched. An unmatched step counts as a plain miss in both scores, so its ordering error is invisible in their comparison.
\item \textbf{Event-conditioned temporal grounding and comprehension.} The post-trained variants are highest on every component: \emph{Matched} has the higher anchor tIoU (0.190) and joint score (0.077), while \reviewhl{\emph{Gold} and \emph{Matched} tie on target tIoU (0.197)} and \emph{Gold} has the higher answer Text F1 and ROUGE-L (0.434 and 0.356). Among off-the-shelf systems, \reviewhl{Qwen3-Omni-30B leads on anchor tIoU} (\reviewhl{0.156}) and joint score (\reviewhl{0.020}), while \reviewhl{InteractiveOmni-4B leads on answer Text F1} (\reviewhl{0.251}) and ROUGE-L (\reviewhl{0.216}). All systems have low anchor and target overlap, and the joint score remains near zero because it requires both grounded spans and a matching answer description. Figure~\ref{fig:test-components} (Right) shows this separation directly: \emph{Gold} attains the highest answer Text F1 (0.434), while its anchor and target tIoU remain low (0.183 and \reviewhl{0.197}). No single component alone establishes complete event-conditioned comprehension.
\end{itemize}

\paragraph{General image and audio evaluation.} A natural concern is that adapting Gemma4-E4B-it toward audio-visual temporal reasoning changes its performance on other multimodal tasks. We evaluate the base model and two post-trained variants on three \emph{out-of-domain} image benchmarks: MMMU-Pro~\citep{yue2025mmmu}, MATH-Vision~\citep{wang2024mathvision}, and OmniDocBench~\citep{ouyang2025omnidocbench}. MMMU-Pro uses its standard split ($n=1{,}730$), MATH-Vision uses its \texttt{test} split ($n=2{,}942$ usable items), and OmniDocBench contains $n=1{,}651$ pages. The two variants are \emph{Gold} and \emph{Matched}, as defined above. For each benchmark we report two complementary signals (Table~\ref{tab:capability-retention}). The first is the task metric: multiple-choice accuracy for MMMU-Pro, answer accuracy across MATH-Vision's multiple-choice and open-ended items, and the official toolkit's text-block normalized edit distance for OmniDocBench. The second is token-weighted micro teacher-forcing perplexity (TF-PPL). With the image or audio, prompt, and a task-specific reference target supplied as one sequence, we mask prompt tokens and report $\exp(\sum_i \mathrm{CE}_i / \sum_i N_i)$, where $\mathrm{CE}_i$ and $N_i$ are the summed cross-entropy and number of scored suffix tokens for item $i$. For MATH-Vision, the target is the solution when available and otherwise the final answer. For OmniDocBench, it is page Markdown assembled from transcribable annotation blocks. The scored suffix also includes the chat template's assistant termination token. Because all variants share an identical processor and chat template, TF-PPL is a parsing-free reference-likelihood probe of distributional shift.
\begin{table*}[t]
\centering
\small
\setlength{\tabcolsep}{4pt}
\renewcommand{\arraystretch}{1.1}
\caption{General image and audio evaluation of the base model and two post-trained variants on eight out-of-domain image/audio benchmarks.}
\label{tab:capability-retention}
\begin{tabular}{@{}lccc@{}}
\toprule
Metric & Gemma4-E4B-it & \shortstack{Temporal (\emph{Gold})} & \shortstack{Temporal (\emph{Matched})} \\
\midrule
\multicolumn{4}{@{}l}{\textbf{Image benchmarks}} \\
MMMU-Pro accuracy $\uparrow$ & \textbf{0.332} & 0.317 & 0.329 \\
MMMU-Pro TF-PPL $\downarrow$ & 17.1 & 14.0 & \textbf{11.2} \\
MATH-Vision answer accuracy $\uparrow$ & 0.125 & 0.124 & \textbf{0.137} \\
MATH-Vision TF-PPL $\downarrow$ & 9.4 & \textbf{7.1} & 7.3 \\
OmniDocBench text edit-dist $\downarrow$ & \textbf{0.682} & 0.704 & 0.702 \\
OmniDocBench TF-PPL $\downarrow$ & 3.873 & 3.573 & \textbf{3.572} \\
\midrule
\multicolumn{4}{@{}l}{\textbf{Audio task subsets}} \\
LibriSpeech-clean WER $\downarrow$ & 0.151 & 0.122 & \textbf{0.091} \\
LibriSpeech-clean TF-PPL $\downarrow$ & 2.587 & 1.937 & \textbf{1.829} \\
CoVoST2 zh$\to$en BLEU $\uparrow$ & \textbf{9.69} & 9.61 & 7.12 \\
CoVoST2 zh$\to$en TF-PPL $\downarrow$ & 42.8 & \textbf{18.1} & 18.6 \\
AudioCaps ROUGE-L $\uparrow$ & 0.072 & 0.078 & \textbf{0.092} \\
AudioCaps TF-PPL $\downarrow$ & 191.8 & \textbf{58.9} & 63.2 \\
OpenHermes-SI ROUGE-L $\uparrow$ & \textbf{0.170} & 0.168 & 0.167 \\
OpenHermes-SI TF-PPL $\downarrow$ & 2.389 & \textbf{2.222} & 2.244 \\
English listening QA (MC) accuracy $\uparrow$ & \textbf{0.899} & 0.876 & 0.885 \\
English listening QA (MC) TF-PPL $\downarrow$ & 59.3 & 32.4 & \textbf{27.2} \\
\bottomrule
\end{tabular}
\end{table*}

We additionally evaluate five audio tasks using processed test subsets from our local AudioLLM v2.1 collection: English ASR from LibriSpeech~\citep{panayotov2015librispeech}, Chinese-to-English speech translation from CoVoST2~\citep{wang2020covost2}, audio captioning from AudioCaps~\citep{kim2019audiocaps}, OpenHermes-derived spoken instruction following, and MCQA from Chinese college-entrance English listening examinations. We use the first 2,000 records in the stored dataset order for each subset, except for spoken instruction following, where all 1,679 records are used. For the audio tasks, we report word error rate (WER) for ASR, corpus BLEU computed with SacreBLEU's 13a tokenizer for translation~\citep{post2018clarity}, ROUGE-L for captioning and spoken instruction following, and option-letter accuracy for listening QA.

On the three evaluated image benchmarks, post-training produces modest changes in task metrics. MMMU-Pro accuracy decreases from $0.332$ to $0.317$--$0.329$, MATH-Vision accuracy ranges from $0.124$ to $0.137$ relative to the base value of $0.125$, and OmniDocBench text-block edit distance increases from $0.682$ to $0.702$--$0.704$. The TF-PPL rows are lower than the base for both post-trained models on all three benchmarks.

The audio results show task-dependent changes. ASR WER decreases from 0.151 for the base model to 0.122 for \emph{Gold} and 0.091 for \emph{Matched}, while AudioCaps ROUGE-L increases from 0.072 to 0.078 and 0.092, respectively. Spoken instruction-following ROUGE-L changes only slightly, from 0.170 to 0.168 and 0.167. Listening-QA accuracy decreases by 2.3 and 1.4 percentage points for \emph{Gold} and \emph{Matched}, respectively. Translation BLEU is similar for the base and \emph{Gold} models (9.69 versus 9.61), but falls to 7.12 for \emph{Matched}. Both post-trained variants have lower TF-PPL on all evaluated image and audio tasks. However, the translation and listening-QA results show that improved reference likelihood does not necessarily translate into improved generation-based task scores. Overall, these evaluations reveal task-specific gains and regressions rather than uniform capability preservation.

\paragraph{Qwen3-Omni-assisted reference construction.} Because Qwen3-Omni-30B is also evaluated, Section~\ref{sec:benchmark} and Table~\ref{tab:provenance} disclose its role in reference construction: it assists Cat1, Cat2, Cat5--Cat7, and limited object-slot filling in Cat4. In Cat2, Qwen3-Omni supports A/V disambiguation and span observations, while Qwen3-235B checks span presence; Cat3 labels are deterministically rendered, and Qwen3-Omni serves only as a speech probe there. Qwen3-Omni-30B leads the off-the-shelf systems on Cat2, Cat5, and Cat6, whereas a post-trained \gemmatemporal{} variant leads on Cat1, Cat3, Cat4, and Cat7. This pattern neither establishes nor rules out same-family construction bias: all systems share frozen silver references, and no human-verified gold subset is available to estimate how construction conventions affect them differently. We therefore interpret its scores with the silver-standard caveats in Section~\ref{sec:benchmark}.

\paragraph{Visual sampling.} Table~\ref{tab:fps-ladder} reports the visual-sampling ladder for Gemma4-E4B-it and the post-trained Temporal (\emph{Gold}) variant on the fixed development suite at rates from 0.125 to 2\,FPS. Gemma4-E4B-it's Cat1, Cat3, and Cat5 scores are highest at 2\,FPS, its Cat7 anchor tIoU is highest at the lowest tested rate, and its Cat2 and Cat6 scores remain low at every tested rate. Temporal (\emph{Gold}) attains its highest Cat1, Cat5, and Cat7 scores at 2\,FPS, ties its highest Cat2 score at 1 and 2\,FPS at the reported precision, but peaks at lower rates on Cat3 (0.125\,FPS), Cat4 (0.25\,FPS), and Cat6 (0.5\,FPS). The performance is non-monotonic over the tested sampling rates, suggesting denser visual sampling does not reliably improve performance across these systems or tasks.

\begin{table*}[t]
\centering
\small
\setlength{\tabcolsep}{3pt}
\renewcommand{\arraystretch}{1.05}
\caption{Visual-sampling results for Gemma4-E4B-it and the post-trained Temporal (\emph{Gold}) variant on the fixed 3,500-item development suite. Each cell reports the category's primary metric. Bold indicates the best value within each model block for each metric column.}
\label{tab:fps-ladder}
\begin{tabular}{@{}lcccccccc@{}}
\toprule
Model & FPS & \shortstack{Cat1\\Acc@1s} & \shortstack{Cat2\\F1@tIoU.5} & \shortstack{Cat3\\Accuracy} & \shortstack{Cat4\\Token F1} & \shortstack{Cat5\\Joint} & \shortstack{Cat6\\Ordered StepF1} & \shortstack{Cat7\\Anchor tIoU} \\
\midrule
\multirow{5}{*}{Gemma4-E4B-it} & 0.125 & 0.084 & \reviewhl{0.019} & 0.438 & \reviewhl{\textbf{0.412}} & 0.109 & \textbf{0.006} & \reviewhl{\textbf{0.098}} \\
 & 0.25 & 0.098 & 0.023 & 0.422 & 0.393 & 0.117 & 0.004 & \reviewhl{0.055} \\
 & 0.5 & 0.200 & 0.020 & 0.430 & \reviewhl{0.385} & 0.113 & 0.002 & \reviewhl{0.041} \\
 & 1.0 & 0.258 & 0.020 & 0.432 & \reviewhl{0.389} & 0.123 & 0.001 & \reviewhl{0.035} \\
 & 2.0 & \textbf{0.264} & \textbf{0.027} & \textbf{0.442} & 0.393 & \textbf{0.146} & 0.002 & \reviewhl{0.048} \\
\midrule
\multirow{5}{*}{\shortstack{Temporal (\emph{Gold})}} & 0.125 & 0.208 & 0.049 & \textbf{0.616} & \reviewhl{0.513} & 0.223 & 0.047 & 0.139 \\
 & 0.25 & 0.252 & 0.061 & 0.574 & \reviewhl{\textbf{0.530}} & 0.267 & 0.034 & 0.143 \\
 & 0.5 & 0.394 & 0.074 & 0.586 & \reviewhl{0.511} & 0.287 & \textbf{0.051} & 0.180 \\
 & 1.0 & 0.468 & \textbf{0.091} & 0.574 & \reviewhl{0.518} & 0.309 & 0.049 & 0.175 \\
 & 2.0 & \textbf{0.472} & \textbf{0.091} & 0.576 & \reviewhl{0.517} & \textbf{0.323} & 0.036 & \textbf{0.184} \\
\bottomrule
\end{tabular}
\end{table*}

\section{Behavioral Perturbations}
\label{sec:probes}

Perturbations are evaluated on the 3,500-item development split. For each category and perturbation, we pair the perturbation result for each item with that same system's full-input result on the same item, and report nonparametric paired-bootstrap 95\% intervals (10,000 resamples) for the mean difference in the category-specific primary metric. All records available in both conditions are retained, with invalid normalized predictions assigned zero on the defined primary metric.

\subsection{Evidence removal, perturbation, and prompting}
\begin{figure*}[t]
\centering
\includegraphics[width=0.98\textwidth]{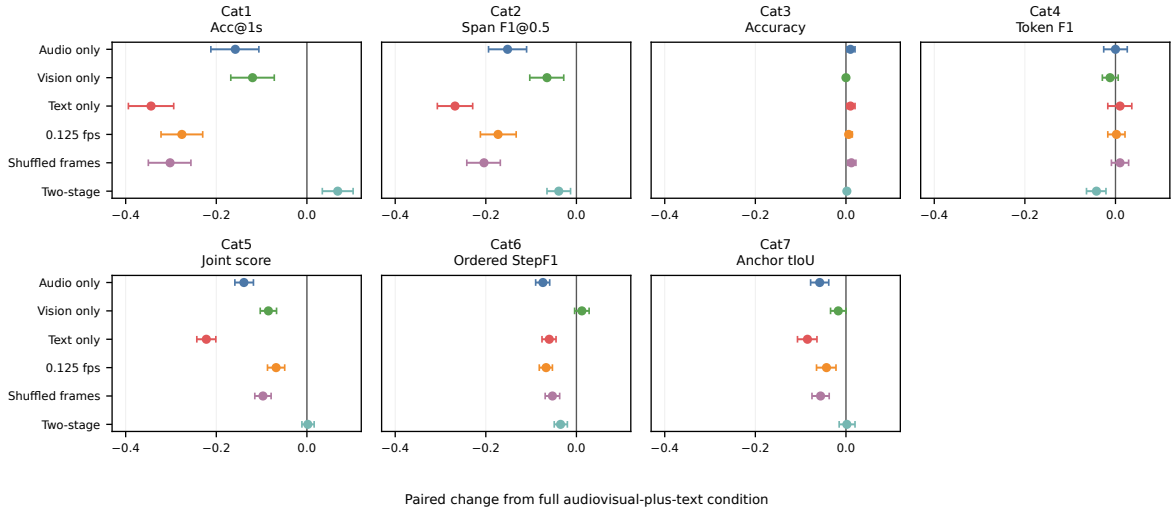}
\caption{Paired development-set effects of evidence removal, visual temporal perturbation, and two-stage prompting for Qwen3-Omni-30B across all seven diagnostic categories. Points show the change in category-specific primary metric relative to the same system's full-input condition. Negative values indicate lower scores under the perturbation. Each estimate uses all 500 category records available in both conditions.}
\label{fig:qwen-probe-effects}
\end{figure*}

Figure~\ref{fig:qwen-probe-effects} summarizes Qwen3-Omni-30B perturbations across all seven categories. Removing an input stream, using a lower visual-sampling configuration, or permuting visual frames is associated with lower development-set scores on several category-specific outcomes, including Cat1 (Minimal-span temporal grounding), Cat2 (Event-span temporal grounding), and Cat5 (Conditional cross-modal localization). Within these categories, the largest displayed decreases occur when the system is given text alone. These effects characterize sensitivity of the evaluated end-to-end system to the available evidence and input interface.

The two-stage condition first elicits a short description of the visual and audible events in the media and their order, then supplies that description with the same media to a second, task-specific generation. Figure~\ref{fig:qwen-probe-effects} shows task-dependent trade-offs rather than a uniform improvement: Cat1 increases by 0.068 (95\% CI [0.034, 0.102]), whereas Cat2, Cat4, and Cat6 decrease by 0.039, 0.042, and 0.035, respectively. Cat5 and Cat7 show no clear change under this  condition. One possible explanation for the Cat1 gain is that the first-pass description provides a compact event-order scaffold for the second-stage timestamp prediction; the mixed effects across categories leave this interpretation open for future investigation.

\section{Discussion}
\label{sec:discussion}

The test results in Table~\ref{tab:test-leaderboard} distinguish reference-text overlap from binding that content to the correct time. Cat4 (Next-step identification) yields nonzero lexical overlap with reference next actions, whereas Cat6 (Temporal chain parsing) exact-chain accuracy is near zero and Cat7 (Event-conditioned temporal grounding and comprehension) grounding overlaps remain low; these metrics differ in scale and difficulty and are not directly comparable across categories, but Cat6 and Cat7 clearly remain challenging for all evaluated systems. The development perturbations show that several scores are sensitive to the availability and ordering of input evidence: Qwen3-Omni-30B loses performance when a lower visual-sampling configuration is used, visual frames are temporally permuted, or visual input is removed. The reference-blind normalizer is intended to reduce the influence of output formatting, but its effect has not been human-verified, so these perturbations do not by themselves separate evidence-use errors from evaluation-pipeline effects. More generally, these perturbations tell us how the evaluated systems behave on this benchmark; they do not reveal the models' internal mechanisms or their capability beyond it. Section~\ref{sec:results} also cannot measure same-family construction bias. Every system is scored against the same frozen references, so these results alone neither establish nor rule out same-family construction bias. A human-verified gold subset remains important for our future work.

\begin{figure*}[t]
\centering
\includegraphics[width=0.94\textwidth]{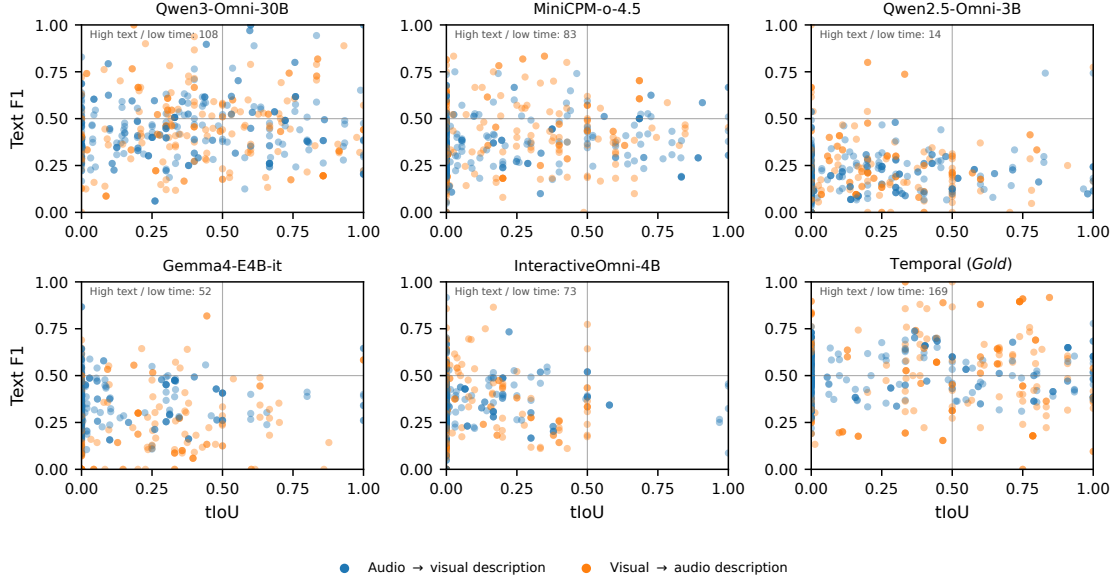}
\caption{Development-set relationship between semantic description and temporal localization for Cat5. Colors indicate the Cat5 construction subtype of each example (audio-to-visual or visual-to-audio description). Vertical and horizontal lines mark tIoU and text-F1 values of 0.5; each panel reports the number of examples with text F1 at least 0.5 but tIoU below 0.5.}
\label{fig:cat5-semantic-temporal}
\end{figure*}

Figure~\ref{fig:cat5-semantic-temporal} illustrates this distinction directly. The Temporal (\textit{Gold}) panel makes it especially clear: 169 development items attain Text F1 at least 0.5 while tIoU is below 0.5, the largest count among the displayed models. Descriptions can therefore overlap substantially with the reference text while localizing the requested cross-modal event poorly in time. Such cases motivate \textbf{joint evaluation of text overlap and temporal localization} rather than treating either metric as a proxy for the other.

Cat3 (A/V synchronization verification) is also unresolved. Despite high valid-output rates, only the two post-trained variants exceed the 0.556 majority-label baseline on the test partition. \emph{Gold}'s affirmative-response rate is 0.364, below the 0.444 reference positive-label rate shown in Figure~\ref{fig:test-components}; these aggregate rates therefore do not establish reliable synchronization discrimination. The reported results are not stratified by desynchronization subtype, so they cannot identify which synchronization cues or rendering types drive these failures.

\section{Related Work}
\label{sec:related}
\paragraph{Temporal diagnostics for video understanding.}
Several benchmarks probe temporal perception in video-language models. TempCompass~\citep{liu2024tempcompass} varies both the temporal aspects under test and the task formats, and constructs conflicting videos that share static content but differ in a specific temporal aspect, preventing models from relying on single-frame or language priors. VITATECS~\citep{li2024vitatecs} introduces a taxonomy of temporal concepts along with counterfactual video descriptions that differ from the original descriptions only in the specified temporal aspect. The Perception Test~\citep{patraucean2023perception} diagnoses broader perceptual, physical, and semantic reasoning skills across video, audio, and text modalities. Whereas TempCompass and VITATECS focus primarily on visual temporal concepts, {\avtrace} evaluates binding audio and vision to a shared physical time axis.

\paragraph{Audio-visual and omnimodal evaluation.} A growing family of benchmarks evaluates joint audio-visual understanding. Daily-Omni~\citep{zhou2025dailyomni} targets cross-modal temporal alignment in question answering. MAVERIX~\citep{xie2025maverix} probes audio-visual integration with multiple-choice and open-ended questions and reports human performance baselines; WorldSense~\citep{hong2025worldsense} evaluates omnimodal understanding with expert-annotated multi-choice QA over audio-synchronized videos. HAVE-Bench~\citep{zhong2026havebench} organizes audio capabilities into a perception--reasoning--interaction hierarchy with multi-turn interaction tasks, and AVHBench~\citep{kim2025avhbench} diagnoses cross-modal hallucination in audio-visual large language models. These benchmarks largely rely on closed-form or descriptive question answering rather than explicit timestamp, span, or ordered-step outputs.

\paragraph{Temporally grounded audio-visual reasoning.} Closer to our work, several efforts require temporally grounded outputs. AVE~\citep{tian2018ave} and OV-AVEBench~\citep{zhou2025ovavel} evaluate audio-visual event localization over short fixed segments. LongVALE~\citep{geng2025longvale} provides large-scale omni-modal event boundary annotations with relation-aware captions for long videos. R-AVST~\citep{zhu2025ravst} evaluates fine-grained spatio-temporal reasoning in complex audio-visual scenarios, and ST-OmniQA~\citep{zeng2026stomniqa} extends evaluation to spatial audio and moving sound sources with temporally grounded reasoning. FAVE~\citep{lu2026fave} benchmarks fine-grained audio-visual temporal perception through cross-modal temporal alignment, event temporal relationships, and moment captioning. {\avtrace} differs from these efforts in scope: it decomposes temporal competence into seven task categories spanning onset grounding, span grounding, synchronization verification, next-action prediction, cross-modal localization, chain parsing, and event-conditioned comprehension. 

\section{Conclusion}

Using {\avtrace}, a seven-task diagnostic benchmark for audio-visual temporal reasoning, we show that the evaluated omni models often produce semantically related text yet obtain low scores on several tasks requiring precise temporal localization, ordered chains, or audio-visual synchronization. Development-set interventions show that several task scores are sensitive to removing or perturbing temporal evidence; these interventions characterize behavioral sensitivity rather than pinpointing its cause. The results motivate models and training objectives that represent event time more explicitly, together with more trustworthy reference data and more comparable evaluation across systems and languages. We will release the benchmark and scoring materials, and hope they serve as a common ground for the community to measure and close the temporal gap in omni models.

\section{Limitations}
\label{sec:limitations}
Our evaluation has limitations in reference quality, input comparability, and language coverage:
\begin{itemize}[leftmargin=*]
  \item {\avtrace} uses silver-standard references. Several categories contain model-assisted annotations or descriptions; the benchmark does not claim human-verified gold correctness or quantify same-family model bias.
  \item Model input interfaces impose different media budgets, sampling policies, and modality representations. Holding these configurations fully constant across systems is often infeasible, so the results characterize end-to-end input configurations rather than matched-compute architectures; this remains a central obstacle to fair cross-model comparison.
  \item The current release does not provide systematic coverage of Southeast Asian languages, cultural contexts, or media. Following our recent SEA-Omni work~\citep{zhang2026unlocking,zhang2026cultural}, we are constructing a Southeast Asia-focused test set with stricter human evaluation, initially supporting Mandarin Chinese, Singapore English, Malay, and Tamil.
\end{itemize}

\section{Acknowledgments}
This research is supported by the National Research Foundation, Singapore under its National
Large Language Models Funding Initiative. Any opinions, findings, conclusions, or recommendations
expressed in this material are those of the author(s) and do not reflect the views of the National
Research Foundation, Singapore. The computational work for this article was fully performed on
resources of the National Supercomputing Centre (NSCC), Singapore (\url{https://www.nscc.sg}).

\bibliography{references}
\bibliographystyle{plainnat}

\appendix

\section*{Appendix A}
\setcounter{subsection}{0}
\renewcommand{\thesubsection}{A.\arabic{subsection}}

\subsection{System input interfaces}
\label{app:systems}
We evaluate each system through its supported input interface. ``Native'' video means the raw media file is passed to the system's own ingestion pipeline; temporal coverage is the portion of each released clip that the system receives under our configuration. All systems decode greedily with a budget of 512 new tokens.
\begin{itemize}[leftmargin=0.4cm]
  \item \textbf{Qwen3-Omni-30B:} Native video configured at \texttt{fps}=1 and \texttt{frames}=4--768, plus the native audio track; full-clip coverage.
  \item \textbf{Qwen2.5-Omni-3B:} Native video with \texttt{fps}=1 and \texttt{max\_frames}=300, sampled uniformly over the released clip, plus native audio with a 300\,s feature window.
  \item \textbf{MiniCPM-o-4.5 (9B):} Native video with \texttt{use\_ffmpeg}=true, \texttt{stack\_frames}=1, and \texttt{max\_slice\_nums}=1, plus the native audio track; full-clip coverage with internal sampling.
  \item \textbf{Gemma4-E4B-it:} Native video clip with \texttt{fps}=1, \texttt{max\_frames}=30, and timestamps formatted as \texttt{mm:ss}, plus extracted 16\,kHz mono WAV audio; coverage of the first 30\,s.
  \item \textbf{InteractiveOmni-4B:} Evaluator-sampled frames with \texttt{fps}=1, \texttt{max\_frames}=30, and \texttt{max\_patch\_num}=4, plus extracted 16\,kHz mono WAV audio; coverage of the first 30\,s.
\end{itemize}

\subsection{Reference construction provenance}
Table~\ref{tab:provenance} is a compact companion to the per-category narrative in Section~\ref{sec:benchmark}, summarizing how each category's released references are constructed. The \emph{Upstream supervision} column states what the source annotations contribute, which is treated as a task-specific signal rather than copied unchanged into the final reference; the two model columns separate the model-assisted stages by role, with Qwen3-Omni-30B performing audio-visual grounding, cross-modal description, and silver-answer construction, and the text-only Qwen3-235B handling question generation and validity checking; and the \emph{Deterministic gates} column lists the code-enforced structural, temporal, and schema checks applied before release. Human review informed pipeline revision across all categories but does not constitute per-record labeling.
\begin{table*}[t]
\centering
\scriptsize
\setlength{\tabcolsep}{4pt}
\renewcommand{\arraystretch}{1.15}
\caption{Reference construction across the seven categories.}
\label{tab:provenance}
\begin{tabular}{@{}>{\raggedright\arraybackslash}p{.04\textwidth}>{\raggedright\arraybackslash}p{.19\textwidth}>{\raggedright\arraybackslash}p{.24\textwidth}>{\raggedright\arraybackslash}p{.19\textwidth}>{\raggedright\arraybackslash}p{.26\textwidth}@{}}
\toprule
Cat. & Upstream supervision & Qwen3-Omni role & Qwen3-235B role & Deterministic gates \\
\midrule
Cat1 & Coarse audio-event windows & Onset detection; clip summary (PT) & Question generation (PT) & Reject out-of-clip onsets; prepend near-start context \\
Cat2 & Event/action intervals & Span observations; A/V disambiguation & Span presence check & Discard absent spans; pad boundary spans \\
Cat3 & Synchronized media & Speech probe only & --- & Rendered labels; offset $\ge$ 500\,ms; yes/no rebalancing \\
Cat4 & Ordered action sequences & Object slot filling (PT) & --- & Target-withholding prefix; source-fixed action types \\
Cat5 & Weak A/V event anchors & Event localization; silver answers & Question gen.; QA validity & Reject abstentions and broad spans \\
Cat6 & Instructional step chains & Per-step relabeling/verification & Question generation & $\ge$2 steps; strictly increasing starts \\
Cat7 & Event spans and questions & Anchor refinement; silver answers & Question gen.; abstention screen & Target rules; temporal-geometry check \\
\bottomrule
\end{tabular}
\end{table*}

\subsection{Source provenance}
Table~\ref{tab:source-provenance} identifies the public source datasets represented
in this release. We retain source attribution while constructing derived task
records, but the underlying sources have heterogeneous versions, access procedures,
and media terms. In particular, several sources distribute annotations while requiring
independent retrieval of public videos; this paper does not claim to redistribute
source media.
\begin{table*}[h]
\centering
\scriptsize
\caption{Public source datasets represented in this release. Categories list
the task converters that use each source; they do not imply that every source record
appears in every partition. Upstream access and licensing terms remain those of the
original releases.}
\label{tab:source-provenance}
\begin{tabular}{p{.26\textwidth}p{.34\textwidth}p{.32\textwidth}}
\toprule
Source & Role in \avtrace{} & Release and access note \\
\midrule
Perception Test~\citep{patraucean2023perception} & Cat1, Cat3--5, Cat7; temporal action/sound annotations and video QA & Official dataset release; CC-BY data and Apache-2.0 code. \\
COIN~\citep{tang2019coin} & Cat2--4, Cat6--7; instructional action segments and chains & Official annotations; source videos are YouTube-hosted. \\
AVE~\citep{tian2018ave} & Cat1--3, Cat5, Cat7; audio-visual events and temporal segments & Official AVE release; source-video availability may vary. \\
LLP/AVVP~\citep{tian2020avvp} & Cat2--3, Cat5, Cat7; weak and dense audio-visual event labels & Official AVVP release; code at \url{https://github.com/YapengTian/AVVP-ECCV20}; media access follows the upstream release. \\
OV-AVEBench~\citep{zhou2025ovavel} & Cat1, Cat3, Cat5, Cat7; open-vocabulary audio-visual event annotations & The upstream work calls the task OV-AVEL and the dataset OV-AVEBench. \\
 R-AVST~\citep{zhu2025ravst} & Cat3, Cat5, Cat7; fine-grained audio-visual spatiotemporal annotations & R-AVST annotations reference videos distributed through the official UnAV-100 release (\url{https://unav100.github.io/}). \\
 UnAV-100~\citep{geng2023unav100} & Cat3; audio-visual event clips & Official release; source-video availability may vary. \\
MUSIC-AVQA~\citep{li2022musicavqa} & Cat3; dynamic audio-visual QA & Official release; versioned upstream annotations are retained locally. \\
WASD~\citep{roxo2025wasd} & Cat3; active-speaker video segments & Official dataset release; use is subject to upstream terms. \\
 AVSBench~\citep{zhou2022avsbench} & Cat3; multi-source audio-visual clips & Official release for audio-visual segmentation research. \\
 EPIC-KITCHENS-100~\citep{damen2022epic} & Cat4, Cat6, Cat7; egocentric action segments and chains & Official dataset release; access is subject to upstream terms. \\
MAVERIX~\citep{xie2025maverix} & Cat7; source video-QA annotations & Official benchmark release; access may require registration. \\
\bottomrule
\end{tabular}
\end{table*}

\subsection{Additional development-set diagnostics}
\begin{figure*}[t]
\centering
\includegraphics[width=0.58\textwidth]{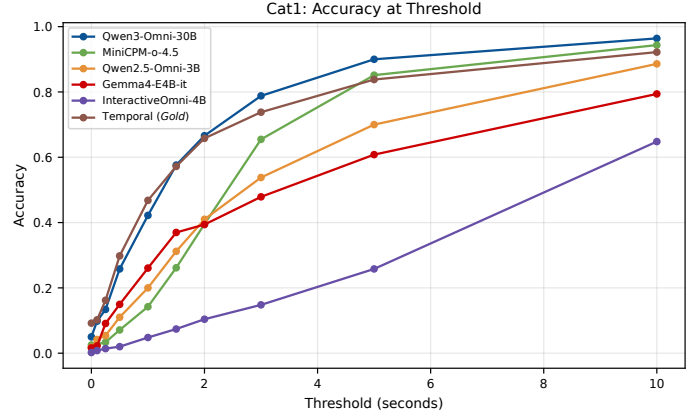}
\caption{Development-set Cat1 accuracy among valid timestamp predictions as the
permitted absolute error increases. The benchmark's primary score remains Acc@1s;
this figure is a sensitivity analysis and does not include invalid outputs.}
\label{fig:appendix-cat1-threshold}
\end{figure*}

The accuracy curves in Figure~\ref{fig:appendix-cat1-threshold} are nondecreasing by construction as the permitted error widens. \emph{Gold} has the highest accuracy through the one-second threshold, while Qwen3-Omni-30B is highest from 1.5 seconds onward; MiniCPM-o-4.5 closes much of the gap by ten seconds. The wide separation at sub-second and one-second tolerances shows that a permissive timestamp threshold can conceal substantial onset-localization error.

\begin{figure*}[t]
\centering
\includegraphics[width=0.98\textwidth]{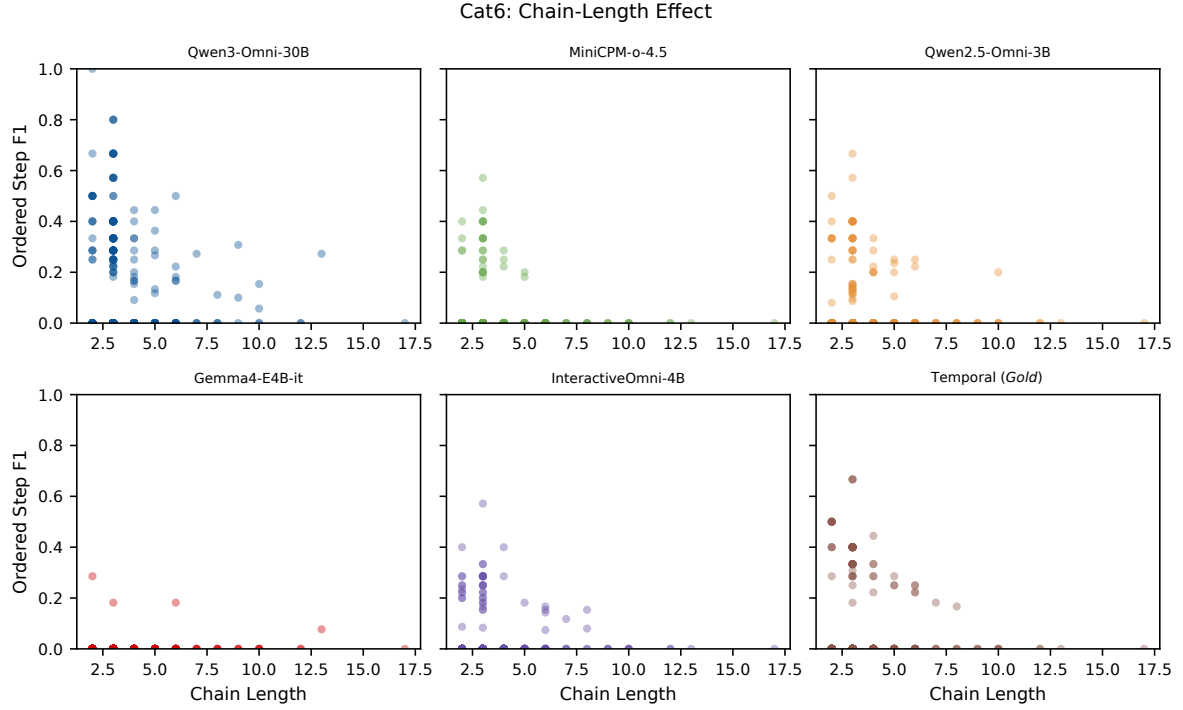}
\caption{Development-set Cat6 ordered StepF1 by reference-chain length.}
\label{fig:appendix-cat6-length}
\end{figure*}

Figure~\ref{fig:appendix-cat6-length} shows that nonzero StepF1 scores are concentrated on shorter reference chains. Qwen3-Omni-30B has the broadest nonzero recovery, but its scores also become sparse as chains lengthen; the other systems are near zero for most longer chains. \emph{Gold} retains nonzero recovery on some two- to eight-step chains, but is zero on all longer chains.

\begin{figure*}[t]
\centering
\includegraphics[width=0.96\textwidth]{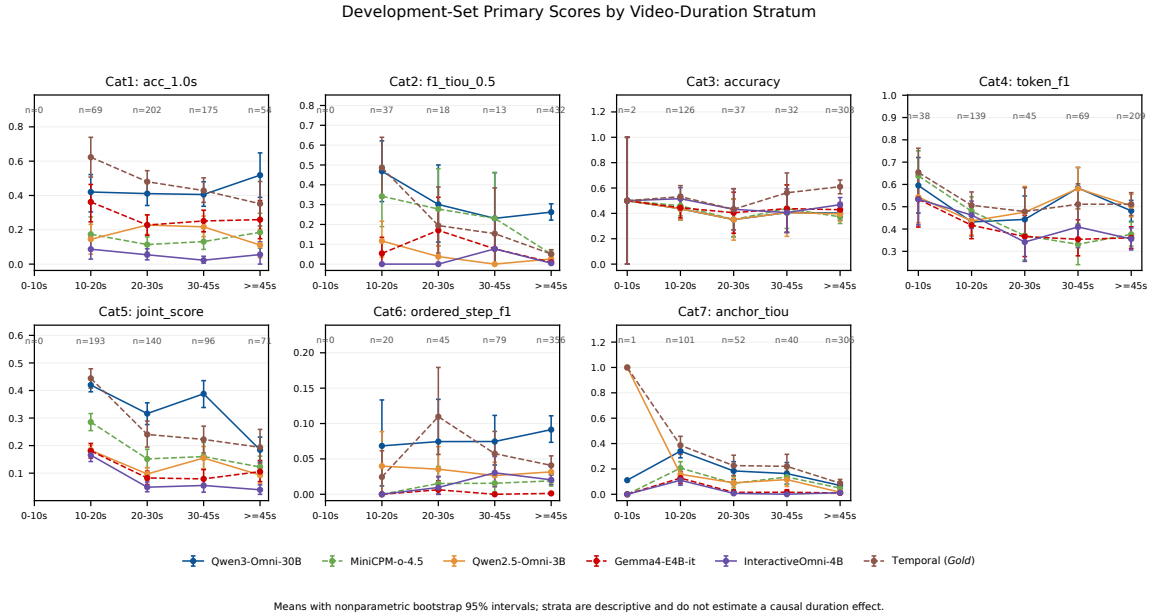}
\caption{Development-set primary scores by video-duration stratum. Each panel uses its category-specific primary metric; points are stratum means and bars are nonparametric bootstrap 95\% intervals. Counts are shared across models within a category, and invalid normalized predictions receive zero on the defined primary metric. The top-coded final stratum contains all videos of at least 45 seconds.}
\label{fig:appendix-duration-strata}
\end{figure*}

Figure~\ref{fig:appendix-duration-strata} shows no uniform duration trend across categories. Cat3, Cat4, and Cat6 vary without a consistent monotonic pattern, whereas Cat2, Cat5, and Cat7 are generally lower in the longest-duration stratum for several systems. Qwen3-Omni-30B is an exception on Cat1, where its score is highest for videos of at least 45 seconds. \emph{Gold} follows the longest-duration decline on Cat2 (0.051 versus 0.486 at 10--20 seconds) and Cat7 (0.088 versus 0.386), whereas its Cat3 accuracy is highest in the longest-duration stratum (0.611). The highly uneven stratum counts, especially for Cat2, mean these differences may also reflect source composition and task difficulty.

\end{document}